# Automated Coding of Communication Data Using ChatGPT: Consistency Across Subgroups

Jiangang Hao, Wenju Cui, Patrick Kyllonen, and Emily Kerzabi

ETS Research Institute, Princeton, NJ 08541, USA

## Abstract

Assessing communication and collaboration at scale depends on a labor-intensive task of coding communication data into categories according to different frameworks. Prior research has established that ChatGPT can be directly instructed with coding rubrics to code the communication data and achieves accuracy comparable to human raters. However, whether the coding from ChatGPT or similar AI technology perform consistently across different demographic groups, such as gender and race, remains unclear. To address this gap, we introduce three checks for evaluating subgroup consistency in LLM-based coding by adapting an existing framework from the automated scoring literature. Using a typical collaborative problem-solving coding framework and data from three types of collaborative tasks, we examine ChatGPT-based coding performance across gender and racial/ethnic groups. Our results show that ChatGPT-based coding perform consistently in the same way as human raters across gender or racial/ethnic groups, demonstrating the possibility of its use in large-scale assessments of collaboration and communication.

## Introduction

The release of ChatGPT in late 2022 marked a turning point in human-AI interaction, demonstrating the ability of large language models (LLMs) to interpret natural language and generate coherent, context-sensitive responses. ChatGPT is a conversational generative AI system that produces responses to user input in a chat format developed by OpenAI. It is powered by underlying LLMs from the GPT family, such as GPT-4 and GPT-4o. This breakthrough fundamentally reshaped access to information, making advanced language technologies widely available to the public and accelerating innovation across domains. Applications now range from customer support systems that provide scalable, real-time assistance (Uzoka et al., 2024) to mental health interventions (Alanezi et al, 2024), as well as emerging uses in education, healthcare, and creative industries. In education, ChatGPT-like generative AI systems are opening new opportunities for both learning and assessment. For learning, these systems can adapt in real time, offering personalized scaffolding across proficiency levels. For instance, Khanmigo, Khan Academy's AI tutor, exemplifies this potential by introducing skilled AI tutors to support personalized learning at scale. For assessment, LLMs enable realistic conversations that support evaluation of complex 21st-century skills such as communication and collaboration (Hao, von Davier et al., 2024).

Collaboration and communication are widely recognized as critical 21st-century skills. Assessing these skills starts with extracting evidence from a large amount of communication data by coding each turn of the communication into categories based on a predefined framework. These codes can indicate the presence or absence of specific collaboration skills (e.g., lack of coded negotiation category may suggest weak negotiation skills), be aggregated into composite scores to capture overall collaboration performance or serve as predictors of task outcomes to differentiate effective from less effective teams (Hao et al., 2019). Traditionally, the coding process is done manually by trained human raters. While effective, this is labor-intensive, time-consuming, and expensive to scale up. Advances in natural language processing (NLP) and machine learning have enabled the development of automated coding systems trained on human-coded data, significantly improving the efficiency and scalability of this process (Flor et al., 2016; Hao et al., 2017a). However, these automated coding systems still depend heavily on large volumes of human-coded data to train the underlying machine learning algorithms. As a result, they only partially address the scalability challenge, rather than fully resolving it.

Rapid advances in LLM-based generative AI provide a promising alternative. Systems such as ChatGPT can be directly instructed with coding frameworks, much like training human raters, to code communication data, thereby reducing the need for extensive human annotation. This consideration is particularly relevant because automated coding is fundamentally a context-based learning task where the LLMs must infer and consistently apply abstract coding criteria from natural-language instructions and a small set of examples, rather than rely on large-scale training data. Despite recent advances, LLMs continue to struggle with complex context-based reasoning and learning, as reflected in the relatively low performance reported on benchmarks such as CL-Bench (Dou et al., 2026). As a result, the effectiveness of LLM-based coding depends on both the capabilities of the underlying LLM and the complexity of the coding task. Early studies showed that LLMs perform well on relatively simple tasks such as sentiment coding (Fatouros et al., 2023). However, their performance on more complex discourse coding remains below expectations, even with state-of-the-art LLMs like GPT-4o (Xu, 2024). As such, a case-by-case and continued evaluation process is needed to identify which coding tasks can be handled at what time by what LLMs.

Hao et al. (2025) demonstrated that ChatGPT can be instructed to code the communication data from collaborative tasks to achieve accuracy comparable to human raters. The findings established the feasibility of coding communication data for typical collaborative problem-solving coding frameworks using top-performing GPT models as of 2025. However, it remains unclear whether such automated coding performs consistently across demographic groups. LLMs are trained on large, web-based corpora. They may inadvertently replicate or amplify biases related to gender, race, or other social categories, as has been demonstrated in applications of automated scoring (Johnson & Zhang, 2024). In this study, we address this gap by empirically examining whether ChatGPT-based coding of communication from collaborative tasks performs consistently in the same way as human raters across demographic groups.

Machine score consistency across subgroups is often discussed under the "fairness" of machine scores in the automated scoring literature (Williamson et al., 2012). But we emphasize that this "fairness" is not the same as score fairness in the psychometric sense, which concerns whether test scores have equivalent meaning, interpretation, and intended use across different groups of examinees (AERA, APA, & NCME, 2014). To better connect with these existing discussions on subgroup comparability of machine scores, we align our discussion to the existing framework after appropriate adjustments due to the nature of the current problem. Williamson et al. (2012) proposed a widely used framework for evaluating subgroup performance of automated scoring systems: (a) examining standardized mean score differences between human and automated scores across subgroups; (b) assessing whether the association between human and automated scores differs by subgroup; (c) evaluating the reliability of automated scores across subgroups; and (d) comparing subgroup differences in predictive validity with respect to a second human rating or an external criterion.

This framework was originally developed for automated scoring systems that assign ordinal or continuous scores to test performance by individual test takers, such as essay scores. In contrast, communication data from collaborative tasks differ in at least two important ways. First, each turn of communication is coded into a nominal category rather than an ordinal score. Although these categorical codes may later be aggregated into ordinal or continuous outcomes (e.g., by counting occurrences of specific categories), the individual codes themselves are usually not scores. Second, the unit of analysis is the communication turn rather than the individual person, and each person may contribute multiple turns of communication, and different numbers of multiple turns. As a result, the mean-based comparison between groups described in (a) of the Williamson et al. (2012) framework is not directly applicable to the coding of communication turns. Accordingly, we introduce three analogous checks for evaluating the performance consistency of LLM-based coding of communication across subgroups. These checks largely correspond to items (b) - (d) of the Williamson et al. framework, with adjustments to accommodate the categorical nature of the coding and the nested structure of the communication data. These checks are operationalized through the following three research questions:

**RQ1:** Is the agreement between ChatGPT-based and human coding consistent across gender and racial/ethnic groups?

**RQ2:** Does the reliability of ChatGPT-based coding differ from that of human coding across gender and racial/ethnic groups?

**RQ3:** Is the subgroup pattern of agreement between ChatGPT-based coding and a secondary human rater comparable to that observed between human raters?

In this paper, we apply ChatGPT to code communication data from three types of collaborative tasks and evaluate the resulting codes using the three checks described above.

Although these checks are slightly narrower in scope than the original Williamson et al. framework, they are well suited to the categorical nature of the coding outcomes and to the nested structure of collaborative interaction data. Beyond the empirical results, the three checks introduced in this study could serve as a basis that may help the research and assessment community establish consensus on subgroup consistency requirements for validating and certifying ChatGPT-based coding. Given the rapid development and growing adoption of LLM-based coding approaches, our findings offer timely evidence on the consistency and reliability of LLM-based coding across subgroups, informing both practical use and ongoing discussions about responsible and fair deployment.

## Literature Review

Collaboration is defined as a "coordinated, synchronous activity that results from a continued attempt to construct and maintain a shared conception of a problem." (Roschelle & Teasley, 1995), which together with communication, is widely recognized as a critical 21st-century skill essential for success in education and the workforce (Fiore et al., 2017; Griffin et al., 2012; OECD, 2017; World Economic Forum, 2025). However, given that these skills are usually displayed in complex interactive scenarios, assessing these skills becomes very challenging and difficult to scale (Hao et al., 2017b).

The advancement of digital technology has enabled computer-mediated online collaboration, making it possible to conduct scale assessments of collaboration, such as the Assessment and Teaching of 21st-Century Skills (ATC21S) (Griffin et al., 2012) and the 2015 Programme for International Student Assessment (PISA 2015) (OECD, 2017). ATC21S assessed collaboration by pairing two students who collaborate via text chat to complete tasks. The scoring was based on students' task responses and actions (Hesse et al., 2015; Scoular et al., 2017). Despite the chat communication containing rich information on collaboration, they were not scored due to technology and cost constraints, limiting the assessment's ability to tap into the key social aspects of collaboration.

PISA 2015 adopted a different approach by placing students in teams with virtual partners programmed to simulate collaboration in a standardized manner (Graesser et al., 2017; He et al., 2017). Students interacted with these virtual agents through predefined text responses designed by experts, ensuring consistency across test administrations. While this approach enhances standardization and reliability, it does not fully reflect the complexities of real-world collaboration, where communication is dynamic and significantly shaped by the unfolding context of prior exchanges. Both ATC21S and PISA 2015 were significantly constrained by technological limitations, psychometric requirements, and the substantial resources needed for implementation (Hao et al., 2019).

A first step in scoring communication and collaboration skills is to code the communication data according to some coding frameworks of interest. Over the past decade, multiple coding frameworks for collaboration have been developed for assessment purposes. The framework from ATC21S defined five core skills essential to effective collaboration: participation, perspective-taking, social regulation, task regulation, and knowledge building (Hesse et al., 2015). The PISA 2025 framework adopted a slightly different approach, focusing on three primary competencies: establishing and maintaining shared understanding, taking appropriate actions to solve problems, and organizing team processes effectively (OECD, 2017). Liu et al. (2016) proposed a framework identifying four key aspects: sharing ideas, negotiating ideas, regulating problem solving, and maintaining communication. Andrews et al. (2017)

introduced a framework that categorizes team interaction patterns, distinguishing between collaborative, cooperative, dominant-dominant, dominant-passive, expert-novice, and even instances of fake collaboration. Andrews-Todd and Kerr (2019) developed a framework that distinguishes between social and cognitive dimensions of collaboration. The social dimension includes maintaining communication, sharing information, establishing shared understanding, and negotiating, while the cognitive dimension encompasses exploring and understanding, representing and formulating, planning, executing, and monitoring. More recently, Kyllonen et al. (2023) developed a framework intended to be applied to different types of collaborative tasks. This framework identifies collaborative problem solving behaviors such as maintaining communication, staying on track, eliciting and sharing information, and acknowledging partners' responses. It emphasizes generalizable interaction patterns and provides a framework to assess CPS across diverse task contexts. Furthermore, the National Assessment of Educational Progress (NAEP) 2026 mathematics framework also integrates key collaborative skills in mathematics, such as attending to and interpreting others' mathematical contributions, evaluating the validity of peers' ideas, and responding constructively to others' reasoning (National Assessment Governing Board, 2023). Although many coding frameworks for collaboration exist, they share considerable overlap, particularly in categories such as idea sharing, idea negotiation, and social communication, with many newer frameworks building upon earlier ones.

Applying these coding frameworks to communication data presents practical challenges. Prior to the development of computer-based NLP technologies, human raters were trained to code these data. This process is labor-intensive and expensive to scale. Advances in NLP and machine learning have enabled automated scoring and coding systems (see Flor & Hao, 2021, for a concise review). The typical workflow usually begins by constructing a training dataset in which human raters code a subset of the communication data (often 10% to 25% of the total dataset, Campbell et al., 2013) based on a coding framework. Each communication turn (e.g., a chat message) in the training data is then transformed into a numerical representation using NLP techniques, such as bag-of-words, n-grams (Jurafsky & Martin, 2019), or more recently, neural embedding vectors (e.g., Devlin et al., 2018; Mikolov et al., 2013; Pennington et al., 2014). Finally, statistical models or supervised machine learning classifiers are trained to learn the relationship between these numerical representations and the coding categories assigned by human raters (Flor et al., 2016; Flor & Andrews-Todd, 2022; Hao et al., 2017b; Moldovan et al., 2011; Rosé et al., 2008). In recent years, the emergence of LLMs has enabled end-to-end approaches for mapping communication data directly to coding categories through fine-tuning the LLM using human-coded training data. Studies have reported that fine-tuned LLMs often outperform traditional machine learning models' coding accuracy (Zhu et al., 2024). While these existing automated coding approaches significantly reduce the need for extensive human coding, the initial development of a training dataset remains labor-intensive and poses a practical barrier to exploring multiple coding frameworks on the same dataset.

Since the release of ChatGPT in late 2022, LLM-based generative AI has shown remarkable progress in aligning LLM responses with human queries. This raises the possibility that we may skip the human coding process altogether by directly instructing LLMs with the coding framework to code data. Figure 1 shows how the ChatGPT-based coding flow is different from a traditional automated coding pipeline. Studies have shown that ChatGPT generally exhibits good accuracy for straightforward coding tasks like sentiment analysis, albeit with some variance across datasets (Belal et al., 2023; Fatouros et al., 2023). However, when confronted with more complex coding tasks, ChatGPT often falls short of meeting expectations (Kocoń et al., 2023; Xu et al., 2024). As such, identifying which coding tasks ChatGPT can reliably and consistently perform has become an important empirical research topic.

**Figure 1**

*Comparison of Traditional and ChatGPT-based Automated Coding*

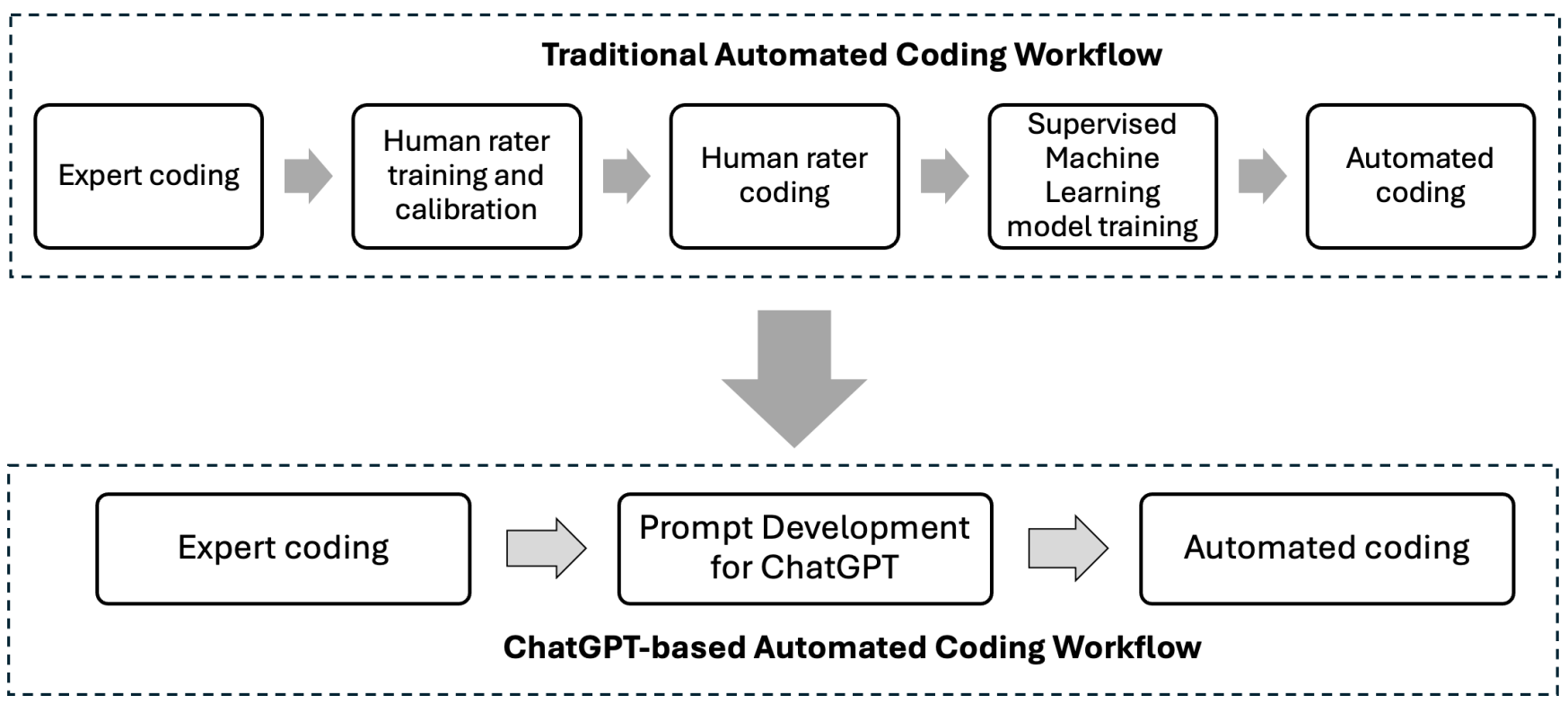

In the context of coding communication data from collaborative tasks, Hao et al. (2025) demonstrated that ChatGPT can be prompted to achieve coding accuracy comparable to that of human raters. However, accuracy alone is not sufficient to warrant its adoption in assessment contexts; coding performance must also be reliable and stable across demographic groups. Because the models underlying ChatGPT (or similar applications) are trained on large-scale internet data with uneven demographic representation, their language understanding capabilities may reflect biases toward dominant linguistic norms, such as those associated with dialectal variation, culturally specific communication styles, or gendered conversational patterns (Blodgett et al., 2020; Fleisig et al., 2024; Kotek et al., 2023). As a result, concerns that LLM-based coding of communication data may exhibit differential performance across subgroups are worth considering.

In the automated scoring literature, consistency of machine scoring performance across subgroups is often discussed under the term "fairness" (e.g., Williamson et al., 2012; Johnson &

McCaffrey, 2023). This notion of fairness, however, differs from that used in psychometrics, which focuses on whether test scores have equivalent meaning, interpretation, and intended use across groups of examinees (AERA, APA, & NCME, 2014). To connect with these existing discussions and account for the specific characteristics of ChatGPT-based coding, this study focuses on the three research questions introduced earlier. The findings are intended to provide empirical evidence on whether ChatGPT-based coding can serve as a scalable alternative or complement to human coding without introducing differential performance across demographic groups.

# Methods

## Datasets

The dataset used in this study was drawn from three collaborative tasks focusing on three general cognitive skills, including *Negotiation*, *Decision-Making*, and *Letter-to-Number* (Kyllonen et al., 2023). Each of these tasks involved a team of four participants collaborating online via text chat. In the *Negotiation* task, team members worked together to plan a fundraising event. Each participant received a list of options, with varying payoffs depending on the individual. Because identical options yielded different payoffs for different participants, the team had to negotiate to reach a mutually agreed solution that maximized individual gains without causing the negotiation to break down, an outcome that would result in no payoff for anyone (Martin-Raugh et al., 2020). In the *Decision-Making* task, four team members engaged in a text-based discussion to select the most suitable apartment from a set of candidates. Each apartment had its own advantages and disadvantages, but each participant only saw a subset of the attributes. To make an informed group decision, team members were to share their unique information with one another. In the *Letter-to-Number* task, participants collaborated to uncover a hidden mapping between letters and numbers, which is a classic reasoning task (Newell & Simon, 1972). Each letter was associated with a predefined number (e.g., A = 1, B = 3, C = 4), and participants proposed operations such as A + B. The system returns the result based on the numerical values of the letters (e.g., A + B = 1 + 3 = 4, so the system responds with C). The goal of the team is to work together to find out the mapping between letters and numbers. Figure 2 presents screenshots of these three tasks.

**Figure 2.**

*Screenshots of the Three Collaborative Tasks for General Cognitive Skill.*

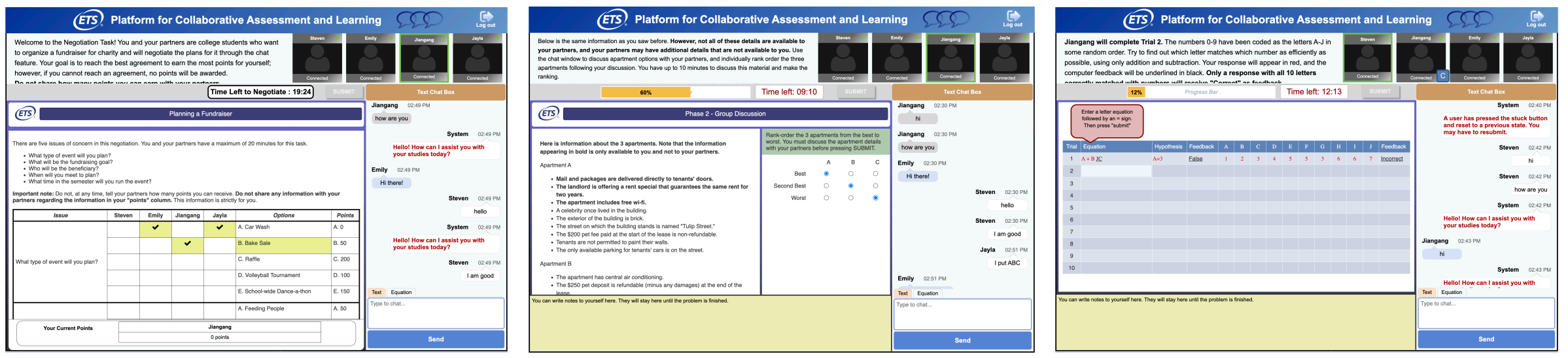

*Note*. Left: *Negotiation* task. Middle: *Decision-Making* task. Right: *Letter-to-Number* task.

Data were collected through a crowdsourcing platform, Prolific (https://www.prolific.com/), and approved by the Institutional Review Board (IRB) of Educational Testing Service. Each collaborative task lasted approximately 40 minutes and generated 50–100 chat turns. The full dataset comprised 8,479 turns of chat across three tasks: *Letter-to-Number* (n = 3,001), *Negotiation* (n = 2,538), and *Decision-Making* (n = 2,940). A total of 431 unique participants contributed, with many participating in more than one task. Of these, 178 self-reported as male (41.3%) and 242 as female (56.2%); 11 participants preferred not to answer. For race, 298 self-reported as White (69.1%), 35 as Black (8.1%), 36 as Hispanic/Latino (8.4%), and 32 as Asian (7.4%); the remaining 30 participants reported multiracial or chose not to report. For the final analyses, we restricted the sample to participants who explicitly identified as either male or female and as White, Black, Hispanic/Latino, or Asian. All other categories, as well as missing responses, were excluded due to small sample sizes.

**Coding Framework**

The coding framework (Kyllonen et al., 2023) used in this study was developed based on the CPS framework from Liu et al. (2017) and a negotiation-specific framework (Martin-Raugh et al., 2020). The same coding framework was applied across all three tasks, with two trained human raters independently coding each chat turn. It is worth noting that for each task, one of the two human raters is considered an expert rater based on expertise and training. In our subsequent comparisons of coding agreement between human raters and AI, we use the expert raters' codes as the reference, though the agreement between AI and the non-expert raters is very similar to that with the expert raters. Table 1 presents the coding categories and their definitions.

**Table 1**

*The CPS Coding Framework Used in this Study*

| |
|---|
| **1. Maintaining communication (MC)**: chats that involve greetings, emotional responses (including emojis), technical discussions, and other communications that cannot be classified elsewhere. |
| **2. Staying on task (OT)**: chats that keep things moving, that involve monitoring time, and steering team effort. |
| **3. Eliciting information (EI)**: chats that elicit information from another about the task, including strategies, goals, and opinions. |
| **4. Sharing information (SI)**: chats that share information, strategies, goals, or opinions. |
| **5. Acknowledging (AK)**: chats involving acknowledging partners' input, stating agreement or accepting a tradeoff, stating disagreement or rejecting a tradeoff, building off one's own or a teammate's idea, and proposing a negotiation tradeoff or suggesting a compromise. |

**LLMs and Prompt Design**

We selected OpenAI's GPT-4o (version: 2024-05-13), deployed via the Azure cloud platform, as the underlying large language model for ChatGPT-based coding in this study. Though GPT-4o is no longer the most powerful GPT model available, its use remains well justified for the present work. GPT-4o represents the final generation of foundational, general-purpose LLMs prior to the introduction of reasoning-oriented models, such as those in the GPT-o series and GPT-5, which augment foundational models with additional multi-step reasoning loops. Importantly, these reasoning-oriented models are built on top of such foundational models, augmenting them with iterative reasoning loops that decompose problems into intermediate steps and refine outputs, thereby improving performance on complex tasks. As such, non-reasoning models like GPT-4o provide a meaningful lower-bound for the performance of LLM-based coding and serve as a baseline for assessing the incremental value of added reasoning capabilities. Moreover, the additional reasoning loops introduced in newer reasoning-oriented models are often optimized for broader deliberative or problem-solving tasks and may not necessarily translate into improved performance for coding tasks. In prior work (Hao et al., 2025), we systematically compared GPT-4o with reasoning-oriented models released through 2025, including GPT-o3, and found that GPT-4o exhibited strong and stable coding performance, underscoring the continued relevance of investigating GPT-4o-like models despite advances in reasoning-based architectures.

LLMs generate text autoregressively by selecting the next token based on the preceding context. At each step, the model produces a vector of logits $z_i$ over the vocabulary, which are converted into probabilities of next token $x_i$ using a temperature-scaled SoftMax: $P(x_i) = \frac{\exp\left(\frac{z_i}{T}\right)}{\sum_j \exp\,(z_j/T)}$, where the temperature $T > 0$ controls the stochasticity of token selection. Lower values of $T$ concentrate probability mass on higher-logit tokens, whereas higher values increase randomness. In standard decoding, tokens are sampled from this distribution, introducing controlled stochasticity that can enhance output diversity, mitigate repetitive or degenerate sequences, and better reflect uncertainty in the model's predictions. However, this stochasticity also introduces more run-to-run variability. While such variability can be beneficial in many generative tasks, it is less desirable in evaluation or benchmark settings where stability and reproducibility are preferred.

In the present study, our objective is to demonstrate whether LLM-based coding can yield consistent performance across demographic subgroups. Therefore, we adopt the greedy decoding, which selects the token with the maximum logit at each step, $x = \arg\max_i z_i$, corresponding to the limiting case of the SoftMax as $T \to 0$. Greedy decoding removes sampling variability, thereby improving the overall reproducibility, even though it does not guarantee strict determinism, as variability may still arise from numerical precision and system-level factors (e.g., Ouyang et al., 2025). It is widely used in evaluation settings to reduce run-to-run variation and improve reproducibility (Renze, 2024; Song et al., 2025).

To optimize performance, we conducted a prompt engineering process following recommended best practices (OpenAI, n.d.). The final prompts were designed to guide the model in coding chat messages accurately without being overly prescriptive and included four components: (1) a statement of the task goal, (2) a description of the coding framework, (3) approximately ten distinct but representative examples per category, and (4) specifications for input and output formats, followed by the actual chat data. Appendix A presents sample prompts. The same prompt structure was used across tasks, with only the task-specific chat examples changed.

**Statistical Analysis**

After assigning both human and AI coding, the resulting data can be cast in a long table format, with columns including Team ID, Person ID, Task ID, AI Coding, Human Coding 1, Human Coding 2, Race, and Gender. By convention, the Human Coding 1 is the expert coding or adjudicated coding from human raters (in those cases in which the two human coders disagreed), which will be used as human coding when comparing with AI coding. Each row of the data table corresponds to a chat turn in the collaboration tasks. As noted earlier, the data have a nested structure: multiple turns are contributed by the same individual, and individuals are further grouped into teams.

To address RQ1, which examines whether agreement between ChatGPT-based and human coding is consistent across gender and racial/ethnic groups, we adopted a generalized linear mixed-effects model (GLMM) for the agreement of human and ChatGPT coding. This modeling approach is analogous to methods commonly used to detect differential item functioning in the measurement literature (Swaminathan & Rogers, 1990). Specifically, we introduce a binary variable $Y$ to represent whether AI and humans agreed on the coding of a given chat turn ($Y = 1$ if they agreed; $Y = 0$ otherwise). Given the nested structure of the data, i.e., multiple chat turns contributed by the same individual, with individuals further nested within teams, a GLMM with a binomial distribution and logit link is used. The fixed effects in the model included the demographic group (gender or race), Task ID, and their interaction. Random effects were included to account for within-person ($u$) and within-team ($v$) dependencies. Specifically, we considered two models: one including the task effect and one without. The model specifications are as follows:

**Model 1**: $\text{Logit}\big(P(Y=1)\big) = \beta_0 + \beta_1 \cdot \text{demographic} + u + v$

**Model 2**: $\text{Logit}\big(P(Y=1)\big) = \beta_0 + \beta_1 \cdot \text{demographic} + \beta_2 \cdot \text{task} + \beta_3(\text{demographic} \cdot \text{task}) + u + v$

Where $u \sim \mathcal{N}(0, \sigma^2_{person})$ and $v \sim \mathcal{N}(0, \sigma^2_{team})$ represent the random intercepts for individuals and teams, respectively. The demographic variable was operationalized as gender (Male vs. Female) or race (Asian, Black, Hispanic, White) depending on the research question being addressed. To align with the DIF analysis used in the discussion of fairness in psychometrics, Male and White were chosen as the reference group, respectively. As a baseline, we also ran the model for the agreement of two human raters. All analyses were conducted in R using the `lme4` package.

To address the RQ2, which examines whether inter-rater reliability between ChatGPT-based and human coding differs from that between human raters across gender and racial/ethnic groups, we computed Cohen's kappa (Cohen, 1965), a widely recognized measure of inter-rater reliability for categorical data. Specifically, Kappa was calculated for human and ChatGPT based coding pairs as well as for pairs of human raters within each demographic group, and these subgroup specific reliability estimates were then compared across groups. To account for the nested structure of the data, we computed 95% confidence intervals for Cohen's kappa using a hierarchical clustered bootstrap. In each bootstrap iteration, teams were sampled with replacement; within each sampled team, persons were sampled with replacement; and all chat turns from each sampled person were retained. Cohen's kappa was then recomputed for each resampled dataset. This procedure was repeated 1,000 times, and the 2.5th and 97.5th percentiles of the bootstrap distribution were used as the 95% confidence interval. To formally compare Human-AI and Human-Human agreement within each task and demographic subgroup, we computed the bootstrap distribution of the difference in kappa, using the same resampled datasets. Two-sided p values were calculated from the centered bootstrap distribution under the null hypothesis that the difference was zero.

To address RQ3, which examines whether the subgroup pattern of agreement between ChatGPT-based coding and a secondary human rater is comparable to that observed between human raters, we followed the same analytic approach as for RQ1, except that the agreement outcome variable $Y$ was defined as agreement between ChatGPT-based coding and the second human rater, and, for comparison, agreement between the two human raters.

## Results

For RQ1, we first examined whether AI coding agreement varied by gender by fitting two GLMMs. In the first model, we included only gender as a fixed effect to examine its impact across all tasks. In the second model, we included gender, task, and their interaction as fixed effects to examine a gender effect at the task level. We chose *Male* as the reference gender and *Letter-to-Number* task as the reference task. The results of the first model show that the effect of gender on human-AI agreement was not significant ($\beta = -0.056$, $SE = 0.062$, $z = -0.90$, $p = .37$), suggesting no overall gender-related bias across all tasks. The second model also showed no significant gender effects or interactions ($ps > .6$). The only significant fixed effect was for the *Negotiation* Task ($\beta = -0.531$, $SE = 0.118$, $z = -4.498$, $p < .001$), indicating lower agreement on the Negotiation task relative to the reference task, consistent with our previous findings (Hao et al., 2025) that the communication contents affected coding performance. Across both models, the random intercepts indicated modest variability at the individual level ($SD \approx 0.216 – 0.217$) and the team level ($SD \approx 0.223 – 0.337$). These estimates suggest that while differences among individuals and teams contribute to variation in AI–human coding agreement, the magnitude of these effects is relatively limited. The detailed model fitting outputs from R are presented in Appendix B.

To examine whether human-AI agreement varied across racial/ethnic groups, we fit two GLMMs in a similar way to the gender analysis, except replacing gender with race. White served as the reference group, and *Letter-to-Number* as the reference task. In the first model, which excluded task effects, none of the racial/ethnic group indicators were significant (ps > .70), indicating no evidence of overall racial/ethnic bias in AI coding across tasks. When we added task and race/ethnicity-by-task interaction terms in the second GLMM, the race/ethnicity effect was not significant. However, there was a significant interaction between being Black and the *Negotiation* task (β = -0.748, SE = 0.317, z = -2.35, p = .018), suggesting that for Black participants, agreement between human and AI ratings was significantly lower in the *Negotiation* task compared to White participants. All other race-by-task interactions were non-significant (ps > .4). When we ran the same analysis for the agreement between human raters, the Black-*Negotiation* interaction was no longer significant. However, a closer look at Cohen's kappa (Table 3) showed that human–AI agreement for Black participants in the *Negotiation* task was comparable to human–human agreement. In contrast, for White participants, the human–AI agreement was much higher than the human–human agreement. This underscores an important point: the apparent race/ethnicity effect stems from a substantial shift in the reference group (White) rather than in the focal group (Black). *Negotiation* task remained a significant predictor of lower agreement across groups (β = -0.434, SE = 0.097, *z* = -4.47, p < .001). Random intercept variances were moderate across models for person (SD = 0.187 – 0.208) and team (SD = 0.226 – 0.351). The R outputs of the model fitting are included in appendix B.

To address RQ2, we report Cohen's kappa in Table 2 and Figure 3 for gender groups and in Table 3 and Figure 4 for racial/ethnic groups. Overall, human-AI and human-human kappa are comparable across both gender and racial/ethnic groups, with some task-specific variation. For example, in the negotiation task (Table 3), the human-AI kappa is significantly higher than the human-human kappa for the White group. Because the human-AI kappa is computed against expert human coding, this suggests that ChatGPT can effectively operationalize the coding framework and produce highly consistent coding in this case. In contrast, for the same task, the human-AI kappa is significantly lower than the human-human kappa for the Asian group. But given the relatively small sample size for the Asian group, this difference is more likely attributable to limited statistical power rather than a systematic performance gap.

**Table 2**

*Cohen's kappa for Different Gender Groups and Tasks*

| Task | Gender | HH | HA | Delta Kappa (HA-HH) (95% C.I.) | *p* | # of Chat Turns |
|---|---|---|---|---|---|---|
| *Letter-to-Number* | Male | 0.747 | 0.724 | -0.023 (-0.092, 0.038) | 0.489 | 1287 |
| | Female | 0.741 | 0.731 | -0.010 (-0.060, 0.043) | 0.715 | 1623 |
| *Negotiation* | Male | 0.491 | 0.565 | 0.074 (-0.021, 0.188) | 0.168 | 1396 |
| | Female | 0.523 | 0.598 | 0.075 (-0.019, 0.165) | 0.104 | 1084 |
| *Decision-Making* | Male | 0.720 | 0.688 | -0.032 (-0.076, 0.011) | 0.128 | 1061 |
| | Female | 0.705 | 0.692 | -0.012 (-0.052, 0.033) | 0.563 | 1830 |
| *All Tasks* | Male | 0.647 | 0.654 | 0.007 (-0.044, 0.066) | 0.761 | 3744 |
| | Female | 0.679 | 0.685 | 0.006 (-0.031, 0.039) | 0.745 | 4537 |

**Note.** HH refers to human-human and HA refers to human-AI (ChatGPT). The C.I.s were computed using hierarchical team and person clustered bootstrapping. The *p* value tests whether Human-AI and human-human kappa differ within each row.

**Figure 3**

*Cohen's kappa for Human–AI and Human–Human Coding Across Gender Groups and Tasks*

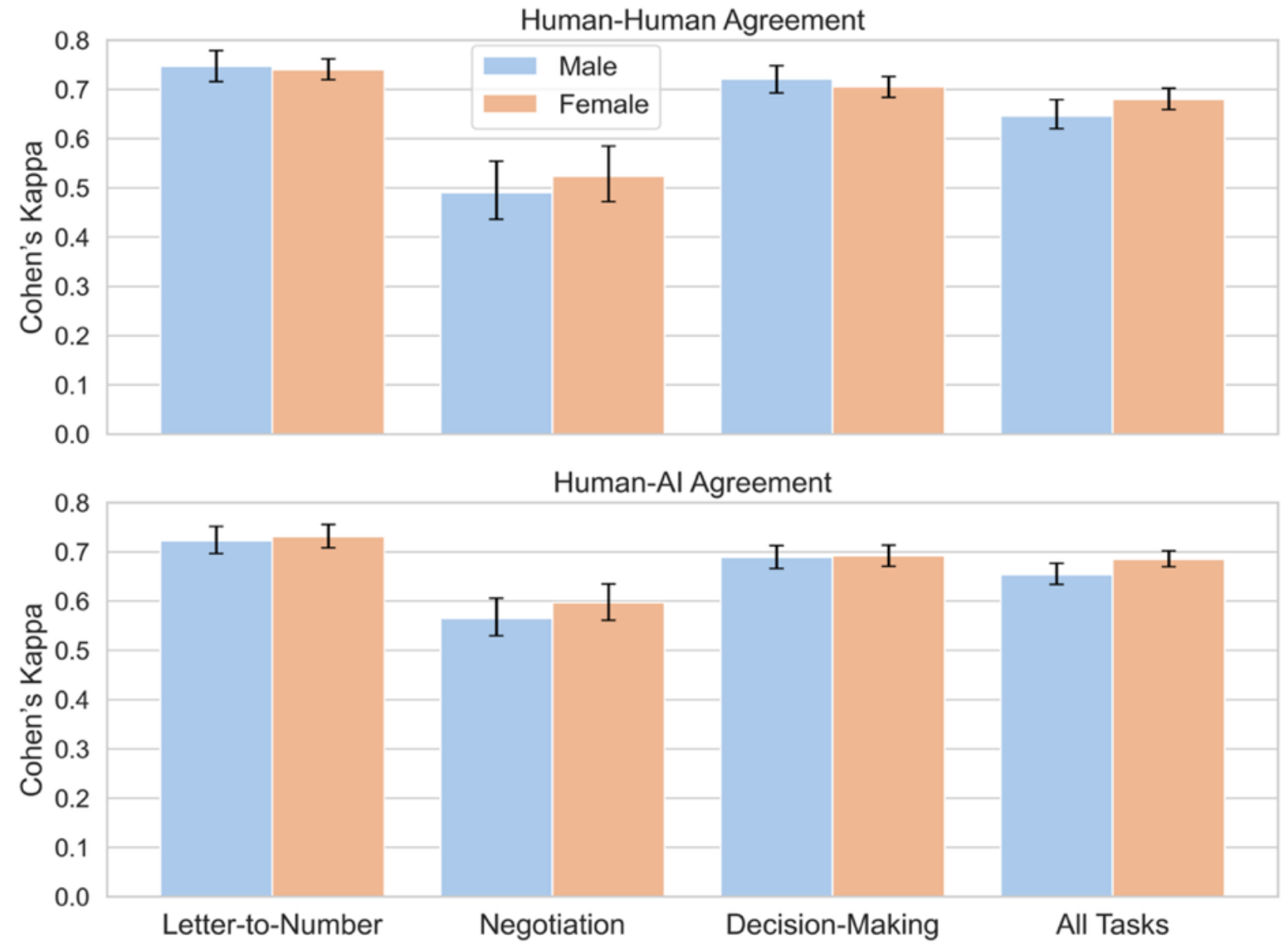

**Note.** The error bars are 95% confidence intervals.

**Table 3**

*Cohen's kappa by Race/Ethnicity and Task*

| Task | Race | HH | HA | Delta kappa (HA–HH) (95% C.I.) | *p* | # of Chat Turns |
|---|---|---|---|---|---|---|
| *Letter-to-Number* | White | 0.728 | 0.726 | -0.003 (-0.056, 0.057) | 0.934 | 2041 |
| | Black | 0.780 | 0.720 | -0.060 (-0.119, 0.010) | 0.066 | 283 |
| | Hispanic | 0.710 | 0.730 | 0.020 (-0.055, 0.100) | 0.607 | 277 |
| | Asian | 0.818 | 0.745 | -0.073 (-0.202, 0.039) | 0.247 | 243 |
| *Negotiation* | White | 0.488 | 0.605 | 0.118 (0.027, 0.220) | 0.013 | 1795 |
| | Black | 0.466 | 0.435 | -0.031 (-0.154, 0.068) | 0.600 | 253 |
| | Hispanic | 0.405 | 0.572 | 0.167 (-0.031, 0.326) | 0.055 | 106 |
| | Asian | 0.634 | 0.476 | -0.158 (-0.272, -0.012) | 0.020 | 176 |
| *Decision-Making* | White | 0.706 | 0.687 | -0.019 (-0.056, 0.018) | 0.293 | 2166 |
| | Black | 0.704 | 0.768 | 0.064 (-0.035, 0.161) | 0.227 | 126 |
| | Hispanic | 0.718 | 0.678 | -0.040 (-0.110, 0.023) | 0.259 | 274 |
| | Asian | 0.739 | 0.728 | -0.011 (-0.095, 0.083) | 0.824 | 255 |
| *All Tasks* | White | 0.653 | 0.678 | 0.025 (-0.015, 0.073) | 0.243 | 6002 |
| | Black | 0.652 | 0.613 | -0.039 (-0.089, 0.017) | 0.156 | 662 |
| | Hispanic | 0.667 | 0.685 | 0.018 (-0.038, 0.081) | 0.555 | 657 |
| | Asian | 0.746 | 0.664 | -0.082 (-0.147, 0.004) | 0.045 | 674 |

**Note.** HH refers to human-human and HA refers to human-AI (ChatGPT). The C.I.s were computed using hierarchical team and person clustered bootstrapping. The *p* value tests whether human-AI and human-human kappa differ within each row.

**Figure 4**

*Cohen's kappa Comparing Human–AI and Human–Human Coding by Race/Ethnicity and Task*

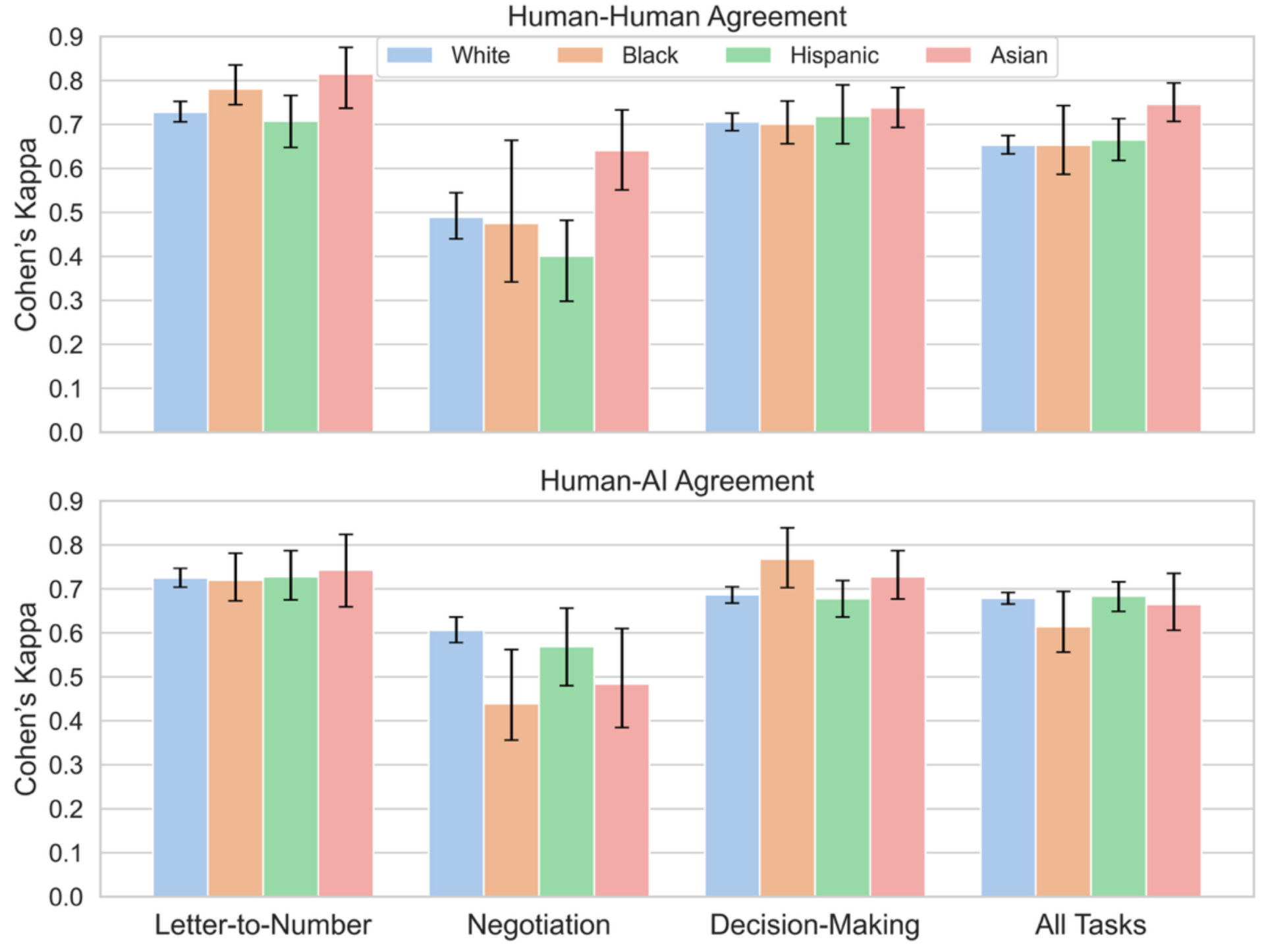

**Note.** The error bars are 95% confidence intervals.

To address RQ3, we examined whether the subgroup pattern of agreement between ChatGPT-based coding and a secondary human rater was comparable to that observed between human raters. Across analyses of both gender and racial/ethnic groups, we found no evidence that agreement with the secondary human rater varied systematically by subgroup for either ChatGPT-based or human coding (see Appendix C for full model outputs). These results also indicate that the relationship between the coding produced by the first human rater or ChatGPT can predict the judgment of the secondary human rater consistently across subgroups. Taken together, these findings suggest that ChatGPT-based coding exhibits subgroup agreement patterns comparable to those observed between human raters within the scope of the current study.

## Discussion

Building on earlier evidence that ChatGPT can achieve coding accuracy comparable to that of human raters, the present study goes beyond accuracy to address a gap in existing evaluation practices by proposing and implementing three subgroup consistency checks for LLM-based coding of communication data. These checks are adapted from a widely used framework originally developed to evaluate subgroup variation in automated scoring of ordinal outcomes and are reformulated to accommodate the nominal nature of communication coding and the interactional structure of the data from collaborative tasks. Specifically, the checks examine (a) whether agreement between ChatGPT-based coding and human coding is consistent across gender and racial/ethnic groups, (b) whether inter-rater reliability between ChatGPT-based and human coding differs from that between human raters across gender and racial/ethnic groups, and (c) whether the subgroup pattern of agreement between ChatGPT-based coding and a secondary human rater is comparable to that observed between human raters. Together, these checks represent an initial step toward a structured approach for evaluating subgroup consistency in LLM-based coding systems and are intended to inform ongoing efforts to build consensus in this area.

We note that a more comprehensive comparison between ChatGPT-based and human coding should extend beyond aggregate agreement measures (e.g., the binary agree/disagree variable or Cohen’s kappa) to examine the distribution of coding categories (e.g., via confusion matrices), which can provide more granular insight into patterns of agreement and disagreement. However, this approach is limited in our setting because the data become sparse when disaggregated to the category level, particularly within subgroups, resulting in unstable estimates and insufficient statistical power to support reliable claims. This challenge is well recognized in psychometric practice, where agreement is often summarized using single indices such as Cohen’s kappa or percent agreement. Given these constraints, we did not include category-level distribution analyses as a primary check, but we acknowledge their value and highlight them as an important direction for future work with larger datasets.

Since most communication data are inherently nested, i.e., multiple turns produced by the same individual, with individuals further grouped within teams, approaches that assume independent observations are not appropriate. To account for dependence among observations from the same person and team, we introduced a GLMM with random intercepts at both the person and team levels. This specification allowed us to account for within-person and within-team dependencies when comparing agreement between AI and human coding across demographic groups. Although modest variability was detected at both levels, the variance components were relatively small, suggesting that clustering by person or team did not strongly influence agreement outcomes. In addition, we complemented the GLMM analyses with Cohen’s kappa, which offers an intuitive measure of agreement across groups and provides further insight into the observed differences.

An important feature of this study is that our analyses drew on three different types of collaborative tasks: *Letter-to-Number*, *Negotiation*, and *Decision-Making*. These tasks differ in structure, cognitive demand, and interaction style, ranging from tightly constrained problem solving (e.g., the *Letter-to-Number* task) to more open-ended group deliberation (e.g., the *Negotiation* task). By demonstrating consistent performance across demographic groups for all three task types, we provide stronger evidence that the observed patterns are not limited to a single task but generalize across diverse collaborative contexts.

Our results provide encouraging evidence that ChatGPT generally codes communication data consistently across the gender and racial/ethnic groups through the lens of the proposed three checks. The only nuance we observed was that in the *Negotiation* task, the AI-human agreement for Black participants appeared lower than that for the White reference group. Importantly, this disparity did not arise because ChatGPT coded chats from Black participants less accurately, but rather because the AI-human agreement for chats from White participants was much higher than the human-human agreement in this task. This elevation in the baseline inflated the relative difference, creating the appearance of a racial/ethnic disparity. In other words, the significant effect reflected a change in the baseline for the White group rather than a systematic bias against Black participants. Several factors may explain why the AI-human agreement was elevated for the White reference group. It is possible that linguistic features or conversational styles more common among White participants aligned more closely with patterns in ChatGPT's training data, leading to high consistency between human and AI coders. Another possibility is that the specific distribution of responses in the White group's Negotiation data happened to match more directly with the coding criteria ChatGPT internalized, resulting in fewer ambiguities compared to human-human coding. Third, chance sampling variation cannot be ruled out, given the modest subgroup sizes. Future studies with large sample size are needed to fully understand the causes. Finally, overall coding agreement for the Negotiation task was lower than that for the other tasks, which may have increased the variability in the coding agreement. This points to an important possibility that reduced coding accuracy may be more likely to introduce less stable performance consistency across demographic groups.

Despite the promising results of this study, several limitations warrant careful considerations. First, the coding performance and consistency achieved with our prompts does not necessarily represent the upper bound; further improvements may be possible through refined prompt design, model fine-tuning, or alternative prompting strategies. Our findings are based on a specific version of GPT (GPT-4o), and LLM capabilities are evolving rapidly. Future iterations of GPT, as well as other LLM families such as Gemini and Claude, may not only deliver improved performance on coding tasks but also introduce variability in performance across demographic groups. Second, the coding frameworks applied here, though typical, were not very complex. Extending LLM-based coding to more complex coding frameworks may present additional challenges and will require further benchmarking to establish both accuracy and consistent performance across subgroups. Third, our findings are conditioned on the available data sample we were able to access, which comprises only a few thousand chat turns. As sample sizes increase, even very small group differences may become statistically significant. This underscores the need for community consensus on what level of group difference should be considered practically meaningful, and the present work aims to contribute to forming such consensus rather than prescribing specific thresholds. Fourth, as noted in the introduction, the coding categories are not themselves the final measures of communication or collaboration but rather the building blocks from which scores are derived. While the individual codes may show no group-wise differences, disparities could still emerge once they are aggregated into composite scores or scales. So, the consistent performance of ChatGPT-based coding does not dictate any fairness derived from these coding, but more a necessary condition. An ultimate discussion of fairness should be at the score and individual or group level, not at the communication turn level. Finally, although ChatGPT and similar systems can apply coding frameworks with accuracy and consistency comparable to human raters, they do not yet fully meet the validity standards expected of human raters (AERA, APA, NCME, 2014; Casabianca et al., 2025). At present, such AI tools are best regarded as a complement to human coding rather than a full replacement, at least until new professional standards and a consensus are established.

Overall, our findings demonstrate that ChatGPT can code communication data accurately and consistently across subgroups, showing strong potential for use in assessment contexts. At the same time, we caution that our results only demonstrate that it *can* do so, but do not provide a warranty that this will hold across all applications. Careful evaluation and benchmarking are therefore essential before deploying it for specific purposes in practice. With appropriate guardrails in place, ChatGPT has the capacity to support scalable assessments of critical 21st-century skills such as collaboration and communication, and AI-driven conversational assessments in general.

## Declaration of generative AI and AI-assisted technologies in the writing process

During the preparation of this work the author(s) used ChatGPT in order to polish the language for clearer presentation. After using this tool/service, the author(s) reviewed and edited the content as needed and take(s) full responsibility for the content of the published article.

## Appendix A: Example Prompts

**Prompts**

Students form a team to work on a computer-based collaborative task. Students use the chat function on the computer to communicate with each other. We will code students' chats into five different categories. The five categories are:

Category 1: Chats that help maintain communication. Such as greeting and pre-task chit-chat, emphatic expression of an emotion or feeling, chat turns that deal with technical issues, and chat turns that don't fit into other coding categories (generally off-topic or typos).
Below is a list of example chats in this category:
["Hi, I'm Jake",
"where are you from?",
"hi, can you see this?",
"lol",
";-)",
"so",
"I'm terrible at this.",
"BOOM",
"Are you a bot?",
"my screen froze",
"my submit button is grayed out",
"who doesn't like discounts, right?"]

Category 2: Chats that focus discussion to foster progress. Such as Making Things Move or Monitoring time to stay on track (not related to task content), steering conversation back to the task.
Below is a list of example chats in this category:
["ah there's 8 minutes",
"let's get this done.",
"hit next",
"we should focus on the task",
"let's talk about the next one",
"what are we supposed to do?",
"We're waiting on Sue to submit",
"I am ok lets move on"]

Category 3: Chats that ask for input on the task. Such as asking for information related to the task, asking for task strategies (regardless of phrasing, no strategy proposed), and asking for task-related goals or opinions. These chats are usually questions and end with a question mark:'?'.
Below is a list of example chats in this category:
["What did you put for best?",
"Do you know what any of the letters are?",

"what about where we'll present?",
"How do I do this?",
"What did you try?",
"Should I add J+J?",
"Why did you put A for best?",
"Why don't you like the raffle?",
"Why are you only adding two numbers?",
"So how about ABC if we all agree?",
"C B A?",
"cba?",
"Can we ...... ?",
"How about ......"]

Category 4: Chats that contribute details for working through the task. Such as sharing information related to the task, sharing strategies for solving the task, and sharing task-related goals or opinions.
Below is a list of example chats in this category:
["Utilities are included in A",
"C=3",
"It's alright, we still know A is 5",
"Let's try adding three numbers",
"If A is 1 then B must be 2",
"If utilities are included, it means the rent is higher",
"I had BCA",
"I want the raffle",
"I think we should do it outside.",
"I don't like paying a pet deposit",
"I'm submitting ACB, since we all agreed.",
"ah, i hate that"]

Category 5: Chats that Acknowledge partner(s) input and may continue off that input with their own. Such as neutrally acknowledge a partner's statement, agree with or support a partner's statement, disagree with a partner's statement, adding details to a previously made chat turn (their own or another player), suggesting a solution of give and take (e.g., if you give me this, I'll give you that. May simultaneously reject an option on the table and propose a compromise).
Below is a list of example chats in this category:
["okay, thanks",
"we nailed it",
"we got all!",
"Good",
"You make a fair point",
"i'll try it",
"Oh I see. That's a good reason.",

"Yeah I wish B had an elevator",
"Alright, same for me",
"weekends are good",
"I don’t think so.",
"can't do afternoons, how about evenings?",
"mine doesn't say that, but I like that A has two bathrooms",
"Good job!"]

Below are the turns of chats. Please assign a category number to each of them and return the coding list. Do not include the original chats.

1. Hello all
2. Hi all!
3. First task, location. we all must agree on a location to "win"
4. I would choose inside due to weather concerns, or virtually. Anyone else have any thoughts?

## Appendix B: GLMM Outputs for RQ1

### Human – AI agreement Across Gender Groups

#### *Not Consider Task Effect*

```
Generalized linear mixed model fit by maximum likelihood (Laplace Approximation) ['glmerMod']
 Family: binomial  ( logit )
Formula: agree_humanai ~ gender + (1 | person_id) + (1 | team_id)
   Data: dff

      AIC       BIC    logLik -2*log(L)  df.resid
   8956.8    8984.9   -4474.4    8948.8      8277

Scaled residuals:
    Min      1Q  Median      3Q     Max
-2.7370  0.3882  0.5037  0.5536  0.9148

Random effects:
 Groups    Name        Variance Std.Dev.
 person_id (Intercept) 0.04723  0.2173
 team_id   (Intercept) 0.11356  0.3370
Number of obs: 8281, groups:  person_id, 538; team_id, 138

Fixed effects:
             Estimate Std. Error z value Pr(>|z|)
(Intercept)   1.23024    0.05494  22.394   <2e-16 ***
genderFemale  0.05591    0.06237   0.896     0.37
---
Signif. codes:  0 ‘***’ 0.001 ‘**’ 0.01 ‘*’ 0.05 ‘.’ 0.1 ‘ ’ 1
```

#### *Consider Task Effect*

```
Generalized linear mixed model fit by maximum likelihood (Laplace Approximation) ['glmerMod']
 Family: binomial  ( logit )
Formula: agree_humanai ~ gender * task_id + (1 | person_id) + (1 | team_id)
   Data: dff

     AIC      BIC   logLik -2*log(L)  df.resid
  8924.6   8980.7  -4454.3    8908.6      8273

Scaled residuals:
    Min      1Q  Median      3Q     Max
-2.5646  0.4037  0.4934  0.5555  0.8985

Random effects:
 Groups    Name        Variance Std.Dev.
 person_id (Intercept) 0.04665  0.2160
 team_id   (Intercept) 0.04950  0.2225
Number of obs: 8281, groups:  person_id, 538; team_id, 138

Fixed effects:
                      Estimate Std. Error z value Pr(>|z|)
(Intercept)           1.341650   0.087567  15.321  < 2e-16 ***
genderFemale          0.008683   0.105480   0.082    0.934
task_idN             -0.531441   0.118139  -4.498 6.85e-06 ***
task_idP              0.127209   0.122636   1.037    0.300
genderFemale:task_idN  0.075078   0.148852   0.504    0.614
genderFemale:task_idP -0.057816   0.150371  -0.384    0.701
---
Signif. codes:  0 ‘***’ 0.001 ‘**’ 0.01 ‘*’ 0.05 ‘.’ 0.1 ‘ ’ 1
```

## Human – AI Agreement Across Racial/Ethnic Groups

### *Not Consider Task Effect*

```
Generalized linear mixed model fit by maximum likelihood (Laplace Approximation) ['glmerMod']
 Family: binomial  ( logit )
Formula: agree_humanai ~ race + (1 | person_id) + (1 | team_id)
   Data: dff
Control: glmerControl(optimizer = "bobyqa", optCtrl = list(maxfun = 1e+07))

     AIC      BIC   logLik -2*log(L)  df.resid
  8637.7   8679.6  -4312.8    8625.7      7989

Scaled residuals:
    Min      1Q  Median      3Q     Max
-2.6480  0.3895  0.5001  0.5532  0.9736

Random effects:
 Groups    Name        Variance Std.Dev.
 person_id (Intercept) 0.04316  0.2078
 team_id   (Intercept) 0.12314  0.3509
Number of obs: 7995, groups:  person_id, 519; team_id, 138

Fixed effects:
             Estimate Std. Error z value Pr(>|z|)
(Intercept)   1.26349    0.04711  26.819   <2e-16 ***
raceAsian     0.01747    0.12145   0.144    0.886
raceBlack    -0.04494    0.11649  -0.386    0.700
raceHispanic  0.02815    0.11668   0.241    0.809
---
Signif. codes:  0 ‘***’ 0.001 ‘**’ 0.01 ‘*’ 0.05 ‘.’ 0.1 ‘ ’ 1
```

### *Consider Task Effect*

```
Generalized linear mixed model fit by maximum likelihood (Laplace Approximation) ['glmerMod']
 Family: binomial  ( logit )
Formula: agree_humanai ~ race * task_id + (1 | person_id) + (1 | team_id)
   Data: dff
Control: glmerControl(optimizer = "bobyqa", optCtrl = list(maxfun = 1e+07))

      AIC       BIC    logLik -2*log(L)  df.resid
   8604.7    8702.5   -4288.4    8576.7      7981

Scaled residuals:
    Min      1Q  Median      3Q     Max
-2.4750  0.4097  0.4926  0.5515  1.0189

Random effects:
 Groups    Name        Variance Std.Dev.
 person_id (Intercept) 0.03505  0.1872
 team_id   (Intercept) 0.05079  0.2254
Number of obs: 7995, groups:  person_id, 519; team_id, 138

Fixed effects:
                      Estimate Std. Error z value Pr(>|z|)
(Intercept)           1.332153   0.063936  20.836  < 2e-16 ***
raceAsian             0.069894   0.182917   0.382   0.7024
raceBlack             0.346599   0.267627   1.295   0.1953
raceHispanic         -0.004952   0.169246  -0.029   0.9767
task_idN             -0.430423   0.096307  -4.469 7.85e-06 ***
task_idP              0.064667   0.093898   0.689   0.4910
raceAsian:task_idN   -0.443741   0.285617  -1.554   0.1203
raceBlack:task_idN   -0.747555   0.317814  -2.352   0.0187 *
raceHispanic:task_idN  0.043851   0.302797   0.145   0.8849
raceAsian:task_idP    0.166875   0.272582   0.612   0.5404
raceBlack:task_idP   -0.244770   0.324222  -0.755   0.4503
raceHispanic:task_idP  0.042882   0.250988   0.171   0.8643
---
Signif. codes:  0 ‘***’ 0.001 ‘**’ 0.01 ‘*’ 0.05 ‘.’ 0.1 ‘ ’ 1
```

# Appendix C: GLMM Outputs for RQ3

## Agreement Across Gender Groups

### *Not Consider Task Effect: Human 1 vs. Human 2*

```
Generalized linear mixed model fit by maximum likelihood (Laplace Approximation) ['glmerMod']
 Family: binomial  ( logit )
Formula: agree_h1h2 ~ gender + (1 | person_id) + (1 | team_id)
   Data: dff

      AIC       BIC    logLik -2*log(L)  df.resid
   8723.2    8751.3   -4357.6    8715.2      8277

Scaled residuals:
    Min      1Q  Median      3Q     Max
-3.1861  0.3280  0.4676  0.5592  1.5341

Random effects:
 Groups    Name        Variance Std.Dev.
 person_id (Intercept) 0.02351  0.1533
 team_id   (Intercept) 0.29837  0.5462
Number of obs: 8281, groups:  person_id, 538; team_id, 138

Fixed effects:
             Estimate Std. Error z value Pr(>|z|)
(Intercept)   1.34103    0.06697  20.024   <2e-16 ***
genderFemale -0.01368    0.06386  -0.214     0.83
---
Signif. codes:  0 '***' 0.001 '**' 0.01 '*' 0.05 '.' 0.1 ' ' 1
```

### *Not Consider Task Effect: AI vs. Human 2*

```
Generalized linear mixed model fit by maximum likelihood (Laplace Approximation) ['glmerMod']
 Family: binomial  ( logit )
Formula: agree_aih2 ~ gender + (1 | person_id) + (1 | team_id)
   Data: dff

      AIC       BIC    logLik -2*log(L)  df.resid
   9580.5    9608.6   -4786.2    9572.5      8277

Scaled residuals:
    Min      1Q  Median      3Q     Max
-2.3429 -1.0021  0.5442  0.6074  1.4459

Random effects:
 Groups    Name        Variance Std.Dev.
 person_id (Intercept) 0.01342  0.1159
 team_id   (Intercept) 0.17459  0.4178
Number of obs: 8281, groups:  person_id, 538; team_id, 138

Fixed effects:
             Estimate Std. Error z value Pr(>|z|)
(Intercept)   1.03816    0.05587  18.580   <2e-16 ***
genderFemale  0.01294    0.05827   0.222    0.824
---
Signif. codes:  0 '***' 0.001 '**' 0.01 '*' 0.05 '.' 0.1 ' ' 1
```

### *Consider Task Effect: Human 1 vs Human 2*

```
Generalized linear mixed model fit by maximum likelihood (Laplace Approximation) ['glmerMod']
 Family: binomial  ( logit )
Formula: agree_h1h2 ~ gender * task_id + (1 | person_id) + (1 | team_id)
   Data: dff

     AIC      BIC   logLik -2*log(L)  df.resid
  8669.2   8725.4  -4326.6   8653.2      8273

Scaled residuals:
    Min      1Q  Median      3Q     Max
-3.0572  0.3497  0.4597  0.5419  1.5075

Random effects:
 Groups    Name        Variance Std.Dev.
 person_id (Intercept) 0.0234   0.1530
 team_id   (Intercept) 0.1443   0.3798
Number of obs: 8281, groups:  person_id, 538; team_id, 138

Fixed effects:
                      Estimate Std. Error z value Pr(>|z|)
(Intercept)            1.56367    0.10061  15.542   <2e-16 ***
genderFemale          -0.09039    0.11198  -0.807    0.420
task_idN              -0.92195    0.13871  -6.647    3e-11 ***
task_idP               0.06435    0.14196   0.453    0.650
genderFemale:task_idN  0.11580    0.15133   0.765    0.444
genderFemale:task_idP -0.02492    0.15861  -0.157    0.875
---
Signif. codes:  0 ‘***’ 0.001 ‘**’ 0.01 ‘*’ 0.05 ‘.’ 0.1 ‘ ’ 1
```

### *Consider Task Effect: AI vs. Human 2*

```
Generalized linear mixed model fit by maximum likelihood (Laplace Approximation) ['glmerMod']
 Family: binomial  ( logit )
Formula: agree_aih2 ~ gender * task_id + (1 | person_id) + (1 | team_id)
   Data: dff

     AIC      BIC   logLik -2*log(L)  df.resid
  9555.9   9612.1  -4770.0   9539.9      8273

Scaled residuals:
    Min      1Q  Median      3Q     Max
-2.2939 -0.9883  0.5386  0.5967  1.4217

Random effects:
 Groups    Name        Variance Std.Dev.
 person_id (Intercept) 0.01365  0.1168
 team_id   (Intercept) 0.11932  0.3454
Number of obs: 8281, groups:  person_id, 538; team_id, 138

Fixed effects:
                      Estimate Std. Error z value Pr(>|z|)
(Intercept)            1.16024    0.09037  12.838  < 2e-16 ***
genderFemale           0.02477    0.10129   0.245    0.807
task_idN              -0.52002    0.12775  -4.071 4.69e-05 ***
task_idP               0.02223    0.12616   0.176    0.860
genderFemale:task_idN -0.08223    0.14107  -0.583    0.560
genderFemale:task_idP -0.03396    0.14174  -0.240    0.811
---
Signif. codes:  0 ‘***’ 0.001 ‘**’ 0.01 ‘*’ 0.05 ‘.’ 0.1 ‘ ’ 1
```

## Agreement Across Racial/Ethnic Groups

## *Not Consider Task Effect: Human 1 vs. Human 2*

```
Generalized linear mixed model fit by maximum likelihood (Laplace Approximation) ['glmerMod']
 Family: binomial  ( logit )
Formula: agree_h1h2 ~ race + (1 | person_id) + (1 | team_id)
   Data: dff
Control: glmerControl(optimizer = "bobyqa", optCtrl = list(maxfun = 1e+07))

      AIC       BIC    logLik -2*log(L)  df.resid
   8463.4    8505.3   -4225.7    8451.4      7989

Scaled residuals:
    Min      1Q  Median      3Q     Max
-3.1140  0.3310  0.4696  0.5571  1.5647

Random effects:
 Groups    Name        Variance Std.Dev.
 person_id (Intercept) 0.02418  0.1555
 team_id   (Intercept) 0.29166  0.5401
Number of obs: 7995, groups:  person_id, 519; team_id, 138

Fixed effects:
              Estimate Std. Error z value Pr(>|z|)
(Intercept)   1.2780171  0.0592974  21.553   <2e-16 ***
raceAsian     0.2090238  0.1321378   1.582    0.114
raceBlack     0.2013260  0.1243620   1.619    0.105
raceHispanic -0.0006173  0.1179456  -0.005    0.996
---
Signif. codes:  0 ‘***’ 0.001 ‘**’ 0.01 ‘*’ 0.05 ‘.’ 0.1 ‘ ’ 1
```

## *Not Consider Task Effect: AI vs. Human 2*

```
Generalized linear mixed model fit by maximum likelihood (Laplace Approximation) ['glmerMod']
 Family: binomial  ( logit )
Formula: agree_aih2 ~ race + (1 | person_id) + (1 | team_id)
   Data: dff
Control: glmerControl(optimizer = "bobyqa", optCtrl = list(maxfun = 1e+07))

      AIC       BIC    logLik -2*log(L)  df.resid
   9230.6    9272.6   -4609.3    9218.6      7989

Scaled residuals:
    Min      1Q  Median      3Q     Max
-2.3594 -1.0176  0.5384  0.6051  1.4292

Random effects:
 Groups    Name        Variance Std.Dev.
 person_id (Intercept) 0.003558 0.05965
 team_id   (Intercept) 0.191258 0.43733
Number of obs: 7995, groups:  person_id, 519; team_id, 138

Fixed effects:
            Estimate Std. Error z value Pr(>|z|)
(Intercept)  1.036505   0.050049  20.710   <2e-16 ***
raceAsian    0.048134   0.111571   0.431    0.666
raceBlack    0.005418   0.108812   0.050    0.960
raceHispanic 0.093959   0.108425   0.867    0.386
---
Signif. codes:  0 ‘***’ 0.001 ‘**’ 0.01 ‘*’ 0.05 ‘.’ 0.1 ‘ ’ 1
```

## *Consider Task Effect: Human 1 vs. Human 2*

```
Generalized linear mixed model fit by maximum likelihood (Laplace Approximation) ['glmerMod']
 Family: binomial  ( logit )
Formula: agree_h1h2 ~ race * task_id + (1 | person_id) + (1 | team_id)
   Data: dff
Control: glmerControl(optimizer = "bobyqa", optCtrl = list(maxfun = 1e+07))

      AIC       BIC    logLik -2*log(L)  df.resid
   8407.7    8505.5   -4189.8    8379.7      7981

Scaled residuals:
    Min      1Q  Median      3Q     Max
-3.3340  0.3161  0.4596  0.5407  1.5220

Random effects:
 Groups    Name        Variance Std.Dev.
 person_id (Intercept) 0.02053  0.1433
 team_id   (Intercept) 0.13532  0.3679
Number of obs: 7995, groups:  person_id, 519; team_id, 138

Fixed effects:
                      Estimate Std. Error z value Pr(>|z|)
(Intercept)           1.496540   0.076408  19.586  < 2e-16 ***
raceAsian             0.007219   0.191417   0.038    0.970
raceBlack            -0.096326   0.258273  -0.373    0.709
raceHispanic          0.089449   0.182651   0.490    0.624
task_idN             -0.916767   0.116618  -7.861  3.8e-15 ***
task_idP             -0.035759   0.111830  -0.320    0.749
raceAsian:task_idN    0.246833   0.309427   0.798    0.425
raceBlack:task_idN    0.330182   0.314096   1.051    0.293
raceHispanic:task_idN -0.328701   0.303093  -1.084    0.278
raceAsian:task_idP    0.496752   0.302161   1.644    0.100
raceBlack:task_idP    0.443277   0.327881   1.352    0.176
raceHispanic:task_idP -0.111246   0.261767  -0.425    0.671
---
Signif. codes:  0 ‘***’ 0.001 ‘**’ 0.01 ‘*’ 0.05 ‘.’ 0.1 ‘ ’ 1
```

## *Consider Task Effect: AI vs. Human 2*

```
Generalized linear mixed model fit by maximum likelihood (Laplace Approximation) ['glmerMod']
 Family: binomial  ( logit )
Formula: agree_aih2 ~ race * task_id + (1 | person_id) + (1 | team_id)
   Data: dff
Control: glmerControl(optimizer = "bobyqa", optCtrl = list(maxfun = 1e+07))

      AIC       BIC    logLik -2*log(L)  df.resid
   9206.4    9304.2   -4589.2    9178.4      7981

Scaled residuals:
    Min      1Q  Median      3Q     Max
-2.3683 -1.0022  0.5345  0.6020  1.5832

Random effects:
 Groups    Name        Variance Std.Dev.
 person_id (Intercept) 0.0000   0.0000
 team_id   (Intercept) 0.1361   0.3689
Number of obs: 7995, groups:  person_id, 519; team_id, 138

Fixed effects:
                      Estimate Std. Error z value Pr(>|z|)
(Intercept)            1.19070    0.07196  16.548  < 2e-16 ***
raceAsian             -0.18227    0.16868  -1.081   0.2799
raceBlack              0.18872    0.25166   0.750   0.4533
raceHispanic           0.05691    0.16335   0.348   0.7276
task_idN              -0.56300    0.11289  -4.987 6.13e-07 ***
task_idP              -0.08265    0.10536  -0.784   0.4328
raceAsian:task_idN     0.30735    0.27239   1.128   0.2592
raceBlack:task_idN    -0.47059    0.29936  -1.572   0.1159
raceHispanic:task_idN  0.01534    0.28951   0.053   0.9577
raceAsian:task_idP     0.43314    0.25427   1.703   0.0885 .
raceBlack:task_idP     0.10661    0.30738   0.347   0.7287
raceHispanic:task_idP  0.08866    0.23895   0.371   0.7106
---
Signif. codes:  0 ‘***’ 0.001 ‘**’ 0.01 ‘*’ 0.05 ‘.’ 0.1 ‘ ’ 1
```